\documentclass[sigconf, authorversion]{acmart}

\def\BibTeX{{\rm B\kern-.05em{\sc i\kern-.025em b}\kern-.08emT\kern-.1667em\lower.7ex\hbox{E}\kern-.125emX}}
\copyrightyear{2019}
\acmYear{2019}
\setcopyright{acmlicensed}

\acmConference{}{}{}

\makeatletter
\renewcommand\@formatdoi[1]{\ignorespaces}
\makeatother

\usepackage{booktabs}
\usepackage{color}

\usepackage{subcaption}
\usepackage{algorithmic}
\usepackage[ruled,linesnumbered]{algorithm2e}

\usepackage{amsmath}
\usepackage{amsfonts}
\usepackage{bbm}
\usepackage{multirow}
\newcommand\mdoubleplus{\mathbin{+\mkern-10mu+}}
\usepackage{amsthm}

\begin{document}
	
	%
	\title{Recurrent Attention Walk for Semi-supervised  Classification }
	
	%

%
%
	
	\author{Uchenna Akujuobi}
	\affiliation{%
		\institution{\textit{King Abdullah University of Science and Technology (KAUST), Saudi Arabia}}
}
	\email{uchenna.akujuobi@kaust.edu.sa}

	\author{Qiannan Zhang}
	\affiliation{%
	\institution{\textit{King Abdullah University of Science and Technology (KAUST), Saudi Arabia}}
	}
	\email{qiannan.zhang@kaust.edu.sa}

	\author{Han Yufei}
	\affiliation{%
		\institution{\textit{Symantec, France}}
	}
		\email{yfhan.hust@gmail.com}
	
	\author{Xiangliang Zhang}
	\affiliation{%
		\institution{\textit{King Abdullah University of Science and Technology (KAUST), Saudi Arabia}}
	}
		\email{xiangliang.zhang@kaust.edu.sa}
	
%
%
%
%
%
%
	%
	\renewcommand{\shortauthors}{Uchenna et al.}
	
	%
	\begin{abstract}
		In this paper, we study the graph-based semi-supervised learning for classifying nodes in attributed networks, where the nodes and edges possess content information. Recent approaches like
		graph convolution networks and attention mechanisms have been
		proposed to ensemble the first-order neighbors and incorporate the
		relevant neighbors. However, it is costly (especially in memory) to consider all neighbors without a prior differentiation. We propose to explore the neighborhood in a reinforcement learning setting and find a walk path well-tuned for classifying the unlabelled target nodes. We
		let an agent (of node classification task) walk over the graph and
		decide where to direct to maximize classification accuracy. We define
		the graph walk as a partially observable Markov decision
		process (POMDP). 
		The proposed method is flexible for working in
		both transductive and inductive setting. Extensive experiments on
		four datasets demonstrate that our proposed method outperforms
		several state-of-the-art methods. Several case studies also illustrate
		the meaningful movement trajectory made by the agent.
	\end{abstract}

	\maketitle
	
	\section{Introduction}
	Network data model interactions between entities such as humans \cite{leskovec2014stanford}, genes \cite{wang2006protein}, and publications \cite{tang2008arnetminer}.
	Networks with node or edge content information are known as \emph{Attributed Networks}. 
	For example, in an attributed web network, nodes are attributed with full website content and edges are attributed with the mention contexts (the sentence encompassing the website mention). 
	A variety of graph mining tasks on attributed networks have been exploited as popular research topics, such as graph embedding \cite{gao2018deep,yang2015network,huang2017label,liao2018attributed}, community detection and clustering \cite{perozzi2014focused,falih2018community}, classification \cite{yang2016revisiting,gcn,velickovic2017graph}, and NLP \cite{gottler1982attributed}. In this paper, we focus on the problem of {semi-supervised node classification on attributed graphs with both nodes and edge contents}. 
	
	\begin{definition}
		\small
		\it
		Semi-supervised Node Classification: Given an attributed graph $G = \{V,E,X_v,X_e\}$, where node set $V$ contains a small subset of labelled nodes $V_l = {<v_i, y_i>},1 \le i \le |V_l|$ and the remaining nodes $ V_u =  V / V_l = {<v_j>}, 1 \le j \le |V_u|$ are unlabeled.   $x_v$ and $x_e$ denote the attributes of nodes and edges in the graph $G$, respectively.  The goal is to infer the labels of the unlabeled nodes $V_u$ based on  the available but limited node labels. Learning from  
		the graph content and structure information.
	\end{definition}
	
	The main solutions to this problem are categorized into two modes: \emph{unsupervised embedding + classifier}, and \emph{semi-supervised learning on graph}. The approaches in  {the first branch} apply a classifier on  {embeddings of graph nodes} learned using   methods like Node2Vec \cite{grover2016node2vec}, DeepWalk \cite{perozzi2014deepwalk}, or TADW \cite{yang2015network}.  {The algorithms belonging to the second branch} directly learn from the graphs, {e.g.,} non-attributed (Label propagation \cite{zhu2002learning} and label spreading  \cite{zhou2004learning}), and attributed graphs embedding (GCN \cite{gcn}, Planetoid \cite{yang2016revisiting}, DGM \cite{akujuobimining}, and \cite{velickovic2017graph,thekumparampil2018attention}).
	{The core ideas behind these approaches} are to 1) jointly learn from the graph structure and the node attributes (most of them are not designed to include edge contents); and 2) aggregate the content of neighboring nodes at different levels of relevance, from immediate neighbors to neighbors $k$-hop away. One limitation of these approaches is 
	the performance downgrade caused by  the noisy information from {an exponentially increasing number of expanded neighborhood members} \cite{zhou2018graph},  {even though} considering high-order structures in graphs might be beneficial for some graph-based problems \cite{lee2018higher,rossi2018estimation,rossi2018higher}. 
	Another issue is the high computational cost, especially in memory cost, caused by the exponentially increasing number of expanded neighbors.  
	
	Furthermore, most of the previously proposed semi-supervised methods are \textbf{transductive}, and thus cannot fit to the situations where new nodes are observed and inserted to the graph. However, deriving embeddings and conducting classification in an inductive way for new unseen nodes is highly demanding in real-world settings, e.g., classifying a new   published paper/website. Inductive approaches also facilitate the generalization across attributed graphs with similar feature spaces   \cite{hamilton2017inductive, yang2016revisiting}. 
	It is thus desirable to design approaches that are flexible for \textbf{both transductive and inductive} setting.
	
	To {\bf reduce the scope of neighbors} to be evaluated in the semi-supervised node classification problem and \textbf{maintain an inductive property},  we propose a recurrent attention framework to learn to explore neighborhoods. In this way we guide neighborhood exploration to better serve the goal of node classification, compared to purely random walk.
	We pose the learn-to-walk task as a partially observable markov decision process (POMDP) problem and attack it with reinforcement learning. 

		To summarize, we address the node classification problem by letting an agent make recurrent decisions on next nodes to visit in its walk on the graph.
		This process can be considered as a recurrent attention-based walk. Therefore, we call our proposed model, \textbf{Recurrent Attention Walk (RAW)}. 
		Comparing to other popular semi-supervised graph-based node classification approaches,  RAW has the following advantages:
		\begin{itemize}
			\item RAW uses a \textbf{recurrent-attention} strategy, while  attention-based node classification approaches like GAT \cite{velickovic2017graph} and AGNN \cite{thekumparampil2018attention}  
			are based on a \textbf{self-attention} strategy, which accessing high-order neighbors by iteratively aggregating one-hop neighbors. By contrast, our \textbf{recurrent-attention} strategy learns how to walk and thus can find the walk path well tuned for classifying the target   nodes,  and thereby minimizing the noisy information obtained. 
			\item RAW thus is more efficient than GCN \cite{gcn} and GAT like approaches on memory cost, because RAW   reduces the number of nodes to aggregate per hop.
			\item RAW is usable in both transductive and inductive settings. We perform extensive experiments on real-world large datasets. The result shows that RAW has superior performance, significantly on inductive setting. 
			\item The walking path generated by RAW can be used to interpret  the decision making process and infer class label dependency, as shown in our case studies. 
		\end{itemize}

	
	\section{Previous Work}
	\label{sec:previous_work}
	In general, solutions for the studied problem (as  defined in the previous section) target on minimizing the loss 
	\begin{equation}\label{eq:e}
	E = \mathcal{E}_l(f(x),y) + \mathcal{E}_r(f(x))\nonumber\\
	\end{equation}
	where $\mathcal{E}_l(f(x),y)$ is the supervised loss function and $\mathcal{E}_r(f(x))$ is the regularizer. 
	The regularizer $\mathcal{E}_r(f(x))$  penalizes a model for assigning different labels $f(x_i) \neq f(x_j)$ to similar nodes $x_i$ and $x_j$, which are close on the graph and have similar content.
	
	Zhu and Ghahramani \cite{zhu2002learning} proposed a transductive label propagation model following the theoretical framework of Gaussian Random Fields to classify nodes in a nearest neighbor graph of a semi-supervised data set. 
	Some other works follow a two-step solution by first learning node embeddings with unsupervised methods  
	\cite{perozzi2014deepwalk,grover2016node2vec,tang2015line}, and then building classifiers on the learned node embedding to infer the unknown labels. 
	Since the embedding is learned in a unsupervised way, it is general enough to be deployed across different tasks (e.g., clustering and link prediction). However, it is not tailored to fit the use in node classification\footnote{We are aware of a big group of related work to our study in graph embedding. In this section, we focus on the most relevant ones solving node classification in semi-supervised learning. Comprehensive discussion of other unsupervised graph embedding for both plain and attributed graphs can be found at \cite{cai2018comprehensive}.}. 
	
	
	Recent decades have witnessed a new trend of research on node classification, which focuses on conducting semi-supervised learning on graphs.
	Yang et al. \cite{yang2016revisiting} proposed a node embedding method to jointly predict the neighborhood context and labels of graph nodes. 
	Kipf et al. \cite{gcn} proposed the use of graph convolutional networks (GCN) for graph-based semi-supervised learning. 
	Zhuang and Ma \cite{zhuang2018dual} extended the idea of GCN by considering global 
	and local 
	consistency. 
	Akujuobi et al. \cite{akujuobimining} studied the use of deep generative models for graph-based semi-supervised learning. 
	Hamilton et al. \cite{hamilton2017inductive} proposed GraphSAGE, an inductive method that computes a node representation by applying an aggregation function over a fixed sample length of node neighbors. 
	
	In general, few of the above-discussed approaches attentively selects the relevant neighboring nodes. The relevance of all neighboring nodes may be implicitly encoded in the aggregation procedure. However, the action on all neighboring nodes without prior preference introduces noisy information due to the exponentially increasing of nodes as the exploration range of the neighborhood extends. To suppress the potential impact of noisy information during aggregating the node neighbors, we propose an attention-based reinforcement learning method for node classification. Next, we survey the use of attention mechanisms and use of reinforcement learning on graph-based problems.
	
	\subsection{Attention-based Node Classification }
	\label{att_learn}
	We can consider selecting the relevant neighboring nodes to visit from the perspective of attention mechanism. Introducing attention mechanism allows the models to focus on the relevant areas of graphs for a given learning task, such as node classification \cite{lee2018attention}.
	Abu-El-Haija et al. \cite{abu2017watch} extends deepwalk by using the attention to guide random walk.
	Thekumparampil et al. \cite{thekumparampil2018attention} 
	introduced attention to the GCN propagation layers to assign more weight to relevant neighbors of each node. 
	Velicovic et al. \cite{velickovic2017graph}, extend the idea of GraphSAGE by introducing the use of attention in the node neighbor sampling.
	Note that the attention neighboorhood per node in the papers, as mentioned above, are the nodes one-hop away from a given node. Our model removes this restriction and thus can achieve better graph exploration. Also, most of these proposed methods do not scale well on large graphs with non-sparse feature vectors as node attributes (i.e., continuous vectors). 
	Furthermore, all these attention models share a \textbf{self-attention} strategy. Specifically, hidden states of each node are computed by attending their neighbors. Thus, by stacking more layers (i.e., $k$-layers), the nodes aggregate information from neighbors up to $k$-hop away. 
	We consider a \textbf{recurrent-attention} strategy, where hidden states of each node are computed by enforcing attention on a recurrent walk on the graph. This strategy reduces the number of nodes to be considered per hop and thereby, minimizing the noisy information obtained. Also, it enables us to evaluate 
	which nodes are more useful based on the information it already gathered from previous hops, and which areas of the graph to explore. 
	
	\subsection{Reinforcement on Graph-Structured Data}
	Several works have studied the application of reinforcement learning on graph-structured data. 
	Hoshen \cite{hoshen2017vain}  applied soft attention on the matrix pair-wise interactions between game agents to select information from relevant agents. 
	Jiang et al. \cite{jiang2018graph}  introduced a graph convolutional reinforcement learning method to learn multi-agent cooperation. 
	Xiong et al. \cite{xiong2017deeppath} proposed a model for finding multi-hop relation paths in knowledge graphs.
	None of these models are designed to select the optimal movement trajectory (path) for node classification. 
	
	The GAM (Graph Attention Model) proposed by Lee et al. \cite{lee2018graph} is an RNN model for graph classification (not node classification), through attention on the graph structural composition. 
	The graph classification differs from node classification on the prediction goal. Thus, the GAM model cannot be applied for node classification as the embedding learned from the graph classification is based on recurrent attention on nodes with random starting nodes. It is not designed to encode a linear combination of the node embeddings. Secondly, the GAM method evaluate the graph label prediction per step iteratively, which is not feasible for node classification in large graphs. GAM also assumes that all the nodes know node types (labels), which does not hold in the settings of semi-supervised node classification.
	
		\begin{figure*}[]
		\centering
		\includegraphics[width=0.9\textwidth]{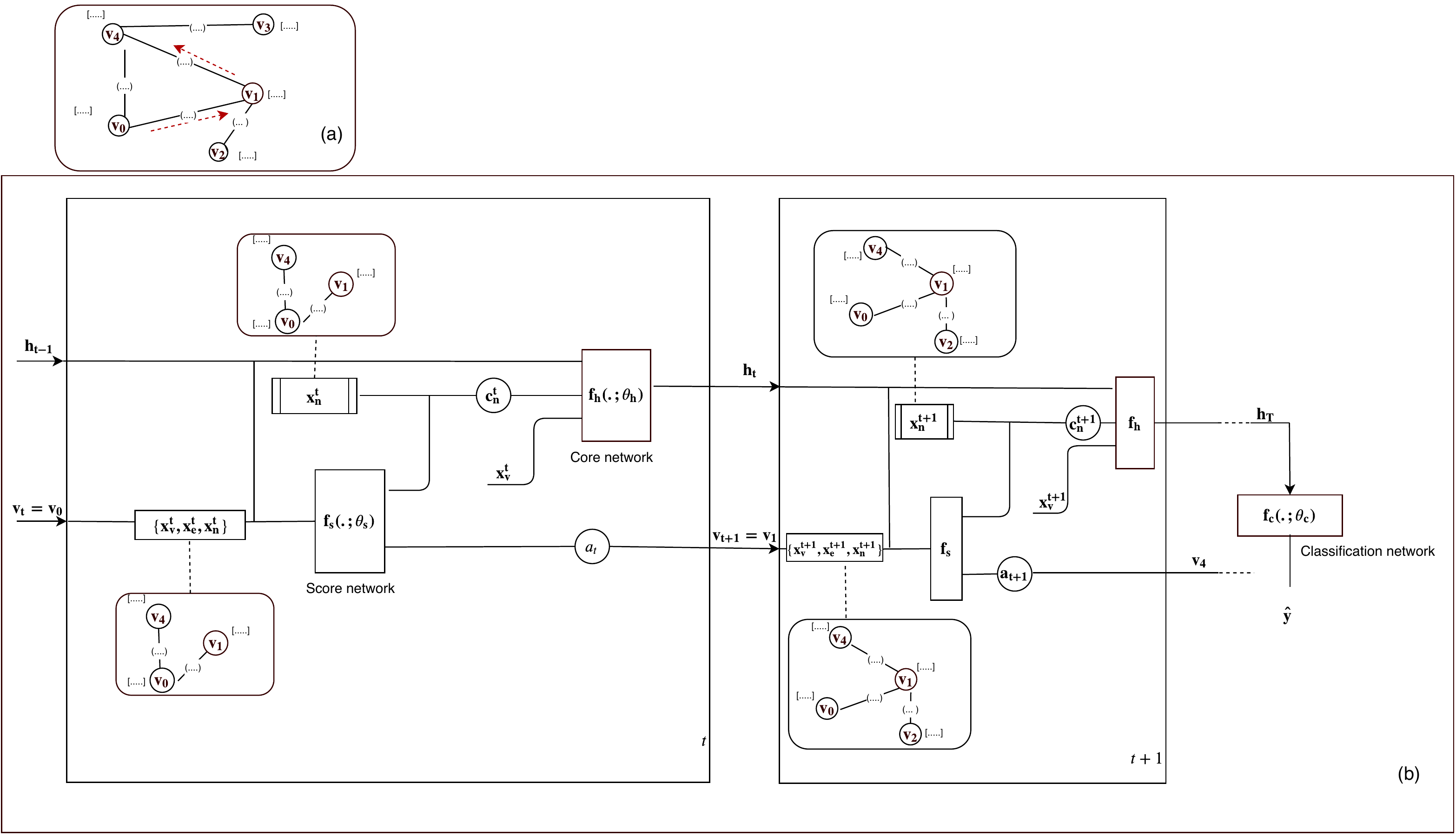}
		\caption{The proposed RAW model. See section \ref{model_description} for description.}
		\label{fig:full_model}
	\end{figure*}
	\section{Methodology}
	
	\subsection{Model Description}
	\label{model_description}
	We model the sequential decision making of which next node to visit by Recurrent Neural Networks (RNN) \cite{cho2014learning} to capture the recurrent dependency in the walk path on the graph. 
	Sequential decision making describes a situation where the decision maker takes its action upon successive observations. The choice of action depends on the expected benefit that can be potentially gained in the future. Given this setting, Markov Decision Process (MDP) provides a coherently appropriate solution to the sequential decision making problem. Nevertheless, exploration of the walk path in an attributed graph violates the Markov property: the observations of the agent at each step should be rich enough to distinguish states of the agent from one to another. In the walk over the graph, observing only the attributes and the neighbors of the current node is not enough to capture all topological information. Therefore, the neighborhood exploration task reduces to a Partially Observable MDP (POMDP) problem. To attack this issue, we encode the past histories of walk paths with RNN to augment state representation of the agent, which facilitates the process of policy learning. 

	As illustrated in Figure \ref{fig:full_model}, the proposed RAW  is composed of 3 networks: the core network, the score network, and the classification network.
	With a small example of an attributed graph in Figure \ref{fig:full_model} (a), the whole process can be explained as follows.
	At the current time $t$,  the agent is   at node $v_0$ and deciding the next visit at time $t+1$, thus $v_t=v_0$.
	In the left of Figure \ref{fig:full_model}  (b),   the score network $f_s(.;\theta_s)$ takes as input the previous history $h_{t-1}$, the current node attribute $x^t_v$, and the attributes of the current neighborhood observation of the agent, which includes the attributes of the immediate node neighbors and edges $\{ x^t_n, x^t_e \}$. The job of the score network is to generate a score for each node neighbor. The generated score in range [0, 1] denotes the relevance of a node neighbor to the given node. After the relevance score is normalized, the next node $v_{t+1}$ to visit is sampled from its neighbors in proportion to their relevance.
	The core network takes over after relevance score is generated. By selectively aggregating the embeddings of  neighboring nodes $x^t_n$ based on the score network, an immediate neighborhood information $c^t_n$ is formed (see section \ref{action} for more details). The core network $f_h(.;\theta_h)$ takes as input the neighborhood aggregation  $c_{n}^t$ , the previous history $h_{t-1}$, and the current node embedding $x^t_v$, and outputs the current walk history $h_t$. This process leads the agent to $v_1$ at time $t+1$, and it repeats to make a move to $v_4$ at $t+2$, etc. After a fixed number of steps $T$,  the \textbf{final} vector $h_T$ summarizing the information obtained from the graph walk is passed to the classification network $f_c(.;\theta_c)$ for the label prediction of the starting node.  See algorithm \ref{alg:generl_alg} for more details.

	RAW is also applicable in  inductive setting, 
	where the walk policy is learned based on the nodes available in the graph. Given a new node added to the graph, the agent initiates a walk from the new unlabelled node guided by the learned policy based on $f_s(.;\theta_s)$ and $f_h(.;\theta_h)$, and finally uses $f_c(.;\theta_c)$ for classification. 

	\begin{algorithm}[hbp!]
		\SetAlgoLined
		\small
		\KwIn{ Graph $G$, start node $v_1$, history vector $h_{0}$ (a vector of zeros), node and edge embeddings $x_v, x_e$}
		\KwResult{label prediction for node $v_1$}
		
		\For{$t \gets 1 \dotsi T$ }{
			Obtain the current node embeddings $x^t_v$ of the current node $v_t$; $x^t_e $ for edges connecting to $v_t$; and $x^t_n$ for neighboring nodes \;
			
			Assign relevance value to each neighbor observation $\varphi^t= f_{s}(h_{t-1}, x^t_v, x^t_e, x^t_n; \theta_s)$\;
			
			Sample next node $v_{t+1}$ from a categorical distribution $Cat (.|P_t(\varphi^t))$ over the neighbors \;
			
			Extract the relevant neighbor information $c^t_n$ \;
			
			update the history vector $h_t = f_h(h_{t-1}, x^t_v, c^t_n;\theta_h)$ \;
			
		}
		
		Obtain the label prediction of the start node $ y_{v_1} = f_c(h_T;\theta_c)$

		\caption{Classifying node $v_1$}
		\label{alg:generl_alg}
	\end{algorithm}
	
	\subsubsection{\textbf{Information Flow}}
	{\sloppy The information flow in RAW has been described above as a sequential decision process, formulated as POMDP. 
		At the time $t$, the agent, which can only observe its one-hop neighbors at the current node, cannot capture the complete topological information in the large graph. Formulating as POMDP allows for a careful treatment to the incomplete observation problem, which is necessary in our case.  

		To address the uncertainty of observation, we  augment  the observation by integrating the information from the previous walk path. This information is encoded recurrently by RNN and updated as the agent traverses. 
	}

	At each step, the agent takes action based on its observation, including the previous history $h_{t-1}$, the current node attribute $x_v^t$, and attributes of its immediate node and edge neighbors, $x_n^t$ and $x_e^t$ respectively, transiting to the next node $v_{t+1}$. 
	The history $h_{t-1}$ acts as a summary of the previous observations in the graph walk, combined with the current observation, the history is updated 
	by the core network $h_t = f_h(h_{t-1}, x^t_v, c^t_n;\theta_h)$, which has GRU at its core and is formulated as:
	\begin{align}
	z_t = \sigma_g (W^z[x_v^t \mdoubleplus c^t_n] + U^zh_{t-1} + b^z),\nonumber\\
	r_t = \sigma_g (W^r[x_v^t \mdoubleplus c^t_n] + U^rh_{t-1} + b^r),\nonumber\\
	h^\prime_t = \sigma_{h^\prime} (W[x_v^t \mdoubleplus c^t_n]+ r_t\circ Uh_{t-1} + b),\nonumber\\
	h_t = z_t \circ h^\prime_t + (1-z_t) \circ h_{t-1}.
	\end{align}
	where $\circ$ and $\mdoubleplus$ denote element-wise multiplication and vector concatenation respectively.  The variable $z_t$ is the update gate which determines the amount of past information to overwrite, $r_t$ is the reset gate which decides the amount of past information to compute a new memory content, $h^\prime_t$ is the current memory content, and $h_t$ is the output vector containing information from the current unit and previous units. The variables $W$ and $U$ are the weights; $x_v^t$ is the node attribute of the current node, $c^t_n$ is the aggregated attribute of the relevant current node neighbors (see section \ref{action}), and $b^z, b^r, b$ are the bias vectors.
	
	At the end of the walk ($t = T$), the core network $f_h(.;\theta_h)$ produces $h_T$, the embedding of the full trajectory started from the target node.
	To classify the target node, $h_T$ is given to  the classification network $f_c(.;\theta_c)$, modeled as a 2-layer neural network, to predict the class label.
	
	\subsubsection{\textbf{Action}}
	\label{action}
	The agent is expected to take actions to   choose the most relevant nodes to visit, and finally collect sufficient information for classifying the target node. 
	Therefore, we can determine the next node to select as an action $a_t$ based on the output $\varphi^t= f_{s}(h_{t-1}, x^t_v, x^t_e, x^t_n; \theta_s)$ of the score network. 
	The output $\varphi^t$ is a measure of relevance between node $v_t$ and its neighbors, and thus, is useful for deciding which of the neighboring nodes are relevant to the current node $v_t$.  
	$\varphi^t$ will be used for the next node selection, and also serve for the history aggregation update. 
	
	The score network is modeled using a sigmoid activation function. Values in  $\varphi^t$ are thus between 0 and 1  for each neighboring node. 
	For the sake of better exploration, a stochastic policy $\pi$ is adopted to make the choice of the next node $v_{t+1}$ to visit via sampling under the categorical distribution $P = Cat(.|\varphi^t)$,  after normalizing $\varphi^t$: 
	$P = Cat(.|\varphi^t) = \frac{1}{\sum_{v_k}  \varphi^t_{v_k}} \times \varphi^t_{v_k}.$
	
	Then the aggregation of relevant neighboring nodes is conducted as:
	\begin{align}
	\label{eq:aggregation}
	c_{n}^t = \sum_{v_k} x_{k} \times  \mathbbm{1}(\varphi^t_{v_k} - 0.5) ; \quad v_k \in N_r(v_t), 
	\end{align}
	where $N_r(v_t)$ is the set of nodes in the one-hope neighborhood of the current node $v_t$, $x_{k}$ is the node attribute of node $v_k$ in the set, and $\varphi^t_{v_k} $ is the relevance score of $v_k$. The indicator function $\mathbbm{1}(.)$ outputs 1 when positive and 0 otherwise.

	\subsubsection{\textbf{Reward}}
	In our model, the performance of a graph walk path (trajectory) would be measured at the end, like evaluating a student passing or failing a course in the final exam after one-semester recurrent study. 
	Specifically, the agent gets an immediate reward  $r_t = 1$  at the last step $T$, if the label prediction at the end ($t=T$)   is correct and $r_t = -1$ otherwise. 
	The goal of the agent is to take actions with large reward to go, $R=\sum^{T}_{t=1} r_t$.
	This reward encourages the agent to explore nodes on the graph that improve the final predictive performance.  
	
	The setting of $T$ is application dependent. A large  $T$ allows for long-run exploration but increases computational cost, while a small $T$ limits the knowledge to aggregate. We have a sensitivity analysis about $T$ in the experimental section. 
	
	\subsection{Training}  
	The final target of our model is to classify an unknown node. Given a trained model, the agent starts from the unknown node, follows the policy to traverse the graph and assigns a label to the given node by the classification network at the end of the graph walk.  To fulfill the goal of the model, it is required to learn a good walk policy and classification network. And we conduct the training process in a semi-supervised manner integrating both labeled nodes $V_l$ and unlabeled ones $V_u$ efficiently.
	
	We augmented the observation to tackle the partial observation problem. But to indicate the property of POMDP, we adopt $o_{1:t}$ to represent the  partial observations along the path until time $t$, while in our study augmented observation $\{h_{t-1}, x_v^t, x_e^t, x_n^t\}$ acts as $o_{1:t}$. We would like to train the policy  $\pi(a_t | o_{1:t};\theta)$ to learn the mapping from the observation space to the action space. Since the policy will take its history from previous transitions as one part of its input, the training of policy will in fact result in an improved core network to provide better history embedding and a score network for more accurate score generation.  Therefore, we train the parameters $\theta = \{\theta_s, \theta_h\}$ together for the policy. The policy objective is the reward in the future over the expectation of the graph walk paths following the current policy, which is  $\mathcal{J}(\theta) = \mathbb{E}_{(\pi; \theta)} \big[\sum_{t=1}^{T} r_t\big]$. 
	
	However, computing the objective function is tough in practice. The expectation over joint probability distribution of walk paths is hard to measure. 
	Therefore, adopting the trick of log derivative to change the gradient of the expectation to the expectation of the gradient, the algorithm REINFORCE for POMDP in \cite{williams1992simple} could take gradients of the objective as following:
	\begin{align}
	\label{rl1}
	\nabla_\theta \mathcal{J} = \sum^{T}_{t=1} \mathbb{E}_{p(o_{1:T}; \theta)}[\nabla_\theta \log \pi (a_t | o_{1 : T}; \theta) R] \nonumber\\
	\quad 
	\approx \frac{1}{M} \sum_{i=1}^{M} \sum^{T - 1}_{t=1} \nabla_\theta \log \pi (a^{i}_t| o^i_{1 : t}; \theta)\gamma^{T-t} R{^i}.
	\end{align}
	The $o^i$'s are the roll-out sequences obtained from running the agent $\pi_{\theta}$  for $i=1,..., M$ episodes, 
	and $\gamma \in (0, 1]$ is a discount factor that gives more preference to actions performed closer to the time the final prediction is made (i.e., $t=T$).  $R^i$ is the reward to go of the  episode $i$. We only adjust the log-probabilities for steps $1 \dotsi T-1$ since there is no choice of next node to visit at time $T$.
	
	On breaking down the joint distribution of the trajectory, the gradient could be estimated by sampling different roll-outs, each running the agent for $T$ limited time steps, from which obtaining the rewards of observations and actions for the estimate. This trial-and-error method is conducted under the current policy, thus providing feedback to the policy and guiding it towards better regions in the parameter space. 
	The information of policy gradients will be back propagated to update parameters of the policy. 
	The differentiable score network $f_s$ and core network $f_h$, represented as neural networks, will be updated. 
	This intuition follows: any gradients of the policy that correspond to high rewards are higher weighted, making roll-outs with higher rewards more likely.

	
	The expected reward of the roll-out only depends on the classification at the end of the walk. Therefore, with the roll-outs  starting at labeled nodes but traversing over unlabelled nodes, the training is  allowed in a semi-supervised manner to use the unlabeled ones as the transitional nodes. It effectively integrates labeled and unlabeled nodes to utilize their information to the maximal extent.
	Besides, high variance from sampling still exists, though the estimate is an unbiased one. The reward setting alleviates this problem in sampled trajectories to some degree by reducing the reward collected at the intermediate steps of roll-outs.

	For the classification $f_c(.;\theta_c)$ network, we define the loss to include the classification error (cross-entropy) and L2 regularization. The classification network is trained in supervised way via gradient descent by itself and provides reward signals to the agent, 
	while score network is trained using REINFORCE. The whole model is trained end-to-end.
	
	\section{Experiments} \label{sec:exp}
	In this section, we present and discuss the extensive experiments and results obtained. We first introduce the four used datasets, the comparison methods, the implementation details, and parameters used for all the models. Finally, we report the results obtained and also present a case study. 
	
	\subsection{Datasets}
	The evaluation datasets are citation networks constructed from Cora, DBLP, and Delve datasets. 
	For each of the resulting paper in the citation networks, we extract the titles (and abstract when available). We also extract the citation context (sentences encompassing the citation) of the references from the papers when available. 
	The statistics of the datasets are shown in Table \ref{dataset_info}. 

	\textbf{CoraL1:} The Cora dataset is extracted from the original Cora data\footnote{https://people.cs.umass.edu/$\sim$mccallum/data.html}. We excluded papers with missing titles and papers with no citation and references (isolated papers). 
	We use the top level labels provided in the dataset.

	\textbf{CoraIDA:} The CoraIDA dataset is constructed as in the CoraL1. However, we only train and test on the papers under Artificial Intelligence, Databases, and Information Retrieval. 

	\textbf{DBLP:} The DBLP dataset was extracted from the DBLP dump\footnote{https://dblp.uni-trier.de/xml/}. This dump is composed of the full DBLP data at the time of download. We extracted papers published in preselected conferences and journals with a focus on predefined topics. Thus, if a paper $X$ is published in one of the database focused conference or journal, paper $X$ is assigned the label ``database''. We constructed a citation network by selecting the neighbors (1 hop away) of each paper. For each of the resulting paper, we extract the title (and abstract when available). This dataset has no edge attribute since the DBLP has no full-text content information.

	\textbf{Delve:} The delve dataset is extracted from the delve website\footnote{http://adatahub.com}. 
	Just as in the DBLP data, we extracted papers published in pre-selected conferences/journals targeting some predefined topics. The citation graph and paper labeling were constructed in the same ways as in DBLP. 

	\begin{table}[]
		\caption{Statistics of datasets used in the evaluations.}
		\centering
		\footnotesize 
		\begin{tabular}{|l|cccc|}
			\hline
			& \# Nodes & \#Edges   & \#labels   & \# Labeled nodes    \\ \hline
			CoraL1   & 31,314     & 133,491   & 10                 & 21,112 \\ \hline
			CoraIDA  & 31,314     & 133,491   & 23          & 9,743    \\ \hline
			DBLP     & 1,037,692 & 7,371,345 & 6                 & 238,350 \\ \hline
			DELVE    & 1,229,280 & 4,322,275 & 7                & 665,495 \\ \hline
		\end{tabular}
		\label{dataset_info}
	\end{table}
	
	\begin{table*}
		\caption{Accuracy results on the citation datasets. The percentage values signify the amount of training data used. }
		\centering
		\footnotesize
		\begin{tabular}{lrrrrr} 
			\toprule
			\multicolumn{6}{c}{CORAL1}          \\ 
			\hline
			& 10\%   & 20\%   & 30\%   & 40\%   & 50\%    \\ 
			\hline
			\multicolumn{6}{c}{Transductive}    \\ 
			\hline
			Random             & 21.1 & 21.1 & 20.6 & 20.5 & 20.8  \\
			Node2Vec           & 72.9 & 75.0 & 75.6 & 75.8 & 75.8  \\
			Deepwalk           & 71.8 & 74.2 & 75.2 & 75.3 & 75.3  \\
			TADW~              & 71.3 & 73.8 & 74.7 & 75.5 & 75.9  \\
			Planetoid-T        & 48.0 & 54.3 & 55.7 & 63.0 & 63.4  \\
			GCN\_MLP           & 73.2 & 76.2 & 77.3 & 77.7 & 78.3  \\
			GCN~               & \textbf{80.3} & \textbf{82.4} & \textbf{83.3} & \textbf{83.8} & \textbf{84.2}  \\
			GCN\_cheb          & \textbf{80.4} & \textbf{81.9} & \textbf{82.9} & \textbf{83.6} & \textbf{84.2}  \\
			GAT                & 75.5 & 75.9 & 76.6 & 76.1 & 76.4  \\
			RAW-T              & \textbf{80.1} & \textbf{81.8} & \textbf{82.4} & \textbf{83.7} & \textbf{84.4}  \\ 
			\hline
			\multicolumn{6}{c}{Inductive}       \\ 
			\hline
			Feature            & 70.9 & 72.5 & 73.8 & 74.0 & 74.8  \\
			GraphSAGE-mean     & 73.4 & 77.0 & 77.3 & 78.7 & 79.6  \\
			GraphSAGE-GCN      & 73.9 & 77.3 & 77.6 & 78.5 & 79.2  \\
			GraphSAGE-maxpool  & 71.0 & 76.1 & 77.4 & 78.5 & 79.7  \\
			GraphSAGE-meanpool & 71.4 & 76.3 & 76.6 & 78.3 & 79.5  \\
			GraphSAGE-LSTM     & 71.0 & 75.7 & 76.1 & 77.5 & 78.9  \\
			Planetoid-I        & 61.9 & 71.0 & 71.8 & 71.5 & 73.5  \\
			FastGCN-importance & 76.3 & 78.5 & 79.8 & 80.9 & 81.6  \\
			FastGCN-uniform    & 75.7 & 78.1 & 79.1 & 80.2 & 81.3  \\
			RAW-I              & \textbf{80.1} & \textbf{81.8} & \textbf{82.4} & \textbf{83.6} & \textbf{84.3}  \\
			\bottomrule
		\end{tabular}
		~
		\begin{tabular}{lrrrrr} 
			\toprule
			\multicolumn{6}{c}{CORAIDA}         \\ 
			\hline
			& 10\%   & 20\%   & 30\%   & 40\%   & 50\%    \\ 
			\hline
			\multicolumn{6}{c}{Transductive}                       \\ 
			\hline
			& 14.8 & 14.3  & 15.2 & 14.8 & 14.7  \\
			& 67.8 & 69.6 & 71.1 & 71.8 & 72.7  \\
			& 66.5 & 69.8 & 71.0 & 72.1 & 72.0  \\
			& 68.4 & 70.7 & 71.9 & 72.8 & 74.1  \\
			& 43.1 & - & - & - & - \\
			& 69.4 & 72.4 & 74.3 & 74.9 & 75.1  \\
			& \textbf{76.4} & \textbf{78.1} & \textbf{79.6} & \textbf{80.2} & \textbf{80.7}  \\
			& \textbf{76.1} & \textbf{78.2} & \textbf{79.6} & \textbf{80.0} & \textbf{80.7} \\
			& 68.4 & 69.3 & 70.6 & 70.8 & 70.6  \\
			& \textbf{76.1} & \textbf{78.0} & \textbf{79.7} & \textbf{79.9} & \textbf{80.6} \\ 
			\hline
			\multicolumn{6}{c}{Inductive}       \\ 
			\hline
			& 66.5 & 69.5 & 71.7 & 71.6 & 72.1  \\
			& 68.1 & 71.7 & 78.0 & 74.6 & 74.9  \\
			& 65.1 & 66.1 & 66.8 & 74.3 & 68.3  \\
			& 73.8 & 68.9 & 75.9 & 79.5 & 76.7  \\
			& 64.9 & 70.1 & 74.7 & 79.3 & 80.3  \\
			& 64.9 & 77.0 & 78.1 & 79.1 & 75.0  \\
			& 57.1 & 62.9 & 65.5 & 67.1 & 67.8  \\
			& 5.3 & 74.5 & 76.8 & 77.6 & 78.4  \\
			& 6.1 & 74.0   & 76.4 & 77.4 & 78.0   \\
			& \textbf{76.1} & \textbf{78.0} & \textbf{79.2} & \textbf{79.9} & \textbf{80.4}  \\
			\bottomrule
		\end{tabular}
		
		\begin{tabular}{lrrrrr} 
			\toprule
			\multicolumn{6}{c}{DBLP}                               \\ 
			\hline
			& 10\%   & 20\%   & 30\%   & 40\%   & 50\%    \\ 
			\hline
			\multicolumn{6}{c}{Transductive}                       \\ 
			\hline
			Random             & 20.6 & 20.6 & 20.5 & 20.6 & 20.6  \\
			Node2Vec           & 78.4 & 78.6 & 78.6 & 78.6 & 78.6  \\
			Deepwalk           & 78.4 & 78.4 & 78.6 & 78.6 & 78.5  \\
			RAW-T               & \textbf{80.9} & \textbf{81.6} & \textbf{81.7} & \textbf{82.0  } & \textbf{82.1}  \\ 
			\hline
			\multicolumn{6}{c}{Inductive}                          \\ 
			\hline
			Feature            & 73.6 & 73.9 & 74.1 & 74.1 & 74.2  \\
			GraphSAGE-mean     & 72.6 & 73.1 & 73.6 & 74.6 & 74.9  \\
			GraphSAGE-GCN      & 76.7 & 77.5 & 77.8 & 78.3 & 78.2  \\
			GraphSAGE-maxpool  & 73.8 & 74.9 & 75.9 & 76.6 & 76.7  \\
			GraphSAGE-meanpool & 73.2 & 74.3 & 74.7 & 75.5 & 75.8  \\
			GraphSAGE-LSTM     & 72.2 & 73.9 & 74.8 & 75.6 & 75.0    \\
			Planetoid-I        & 73.8 & 74.6 & 74.5 & 74.6 & 74.9  \\
			FastGCN-importance & 76.4 & 77.9 & 78.8 & 79.2 & 79.3  \\
			FastGCN-uniform    & 76.4 & 77.7 & 78.3 & 78.3 & 78.5 \\
			RAW-I               & \textbf{80.9} & \textbf{81.5} & \textbf{81.6} & \textbf{81.9} & \textbf{82.1}  \\
			\bottomrule
		\end{tabular}
		~
		\begin{tabular}{lrrrrr} 
			\toprule
			\multicolumn{6}{c}{DELVE}                              \\ 
			\hline
			& 10\%  & 20\%   & 30\%   & 40\%   & 50\%    \\ 
			\hline
			\multicolumn{6}{c}{Transductive}                       \\ 
			\hline
			& 24.8 & 24.8 & 24.8 & 24.8 & 24.7  \\
			& 58.7 & 58.8 & 58.8 & 58.8 & 58.8  \\
			& 58.6 & 58.8 & 58.8 & 58.8 & 58.8  \\
			& \textbf{81.6} &\textbf{82.9} & \textbf{83.5} & \textbf{83.6}  & \textbf{84.2}    \\ 
			\hline
			\multicolumn{6}{c}{Inductive}                          \\ 
			\hline
			& 80.8 & 81.0   & 81   & 81.1 & 81.1  \\
			& 74.4 & 76.8 & 78   & 79.1 & 79.8  \\
			& 65.1 & 66.1 & 66.8 & 67.7 & 68.3  \\
			& 74.8 & 77.8 & 78.8 & 79.5 & 80.4  \\
			& 75.2 & 77.4 & 78.4 & 79.3 & 80.3  \\
			& 74.5 & 77.0   & 78.1 & 79.1 & 80.2  \\
			& 78.8 & 78.7 & 79.5 & 79.5 & 79.6  \\
			& 75.5 & 75.7 & 74.9 & 74.3 & 73.9  \\
			& 75.1 & 75.1 & 74.1 & 73.1 & 72.7 \\
			&  \textbf{81.5} &  \textbf{82.8} & \textbf{83.5} &  \textbf{83.6} & \textbf{84.0}     \\
			\bottomrule
		\end{tabular}
	\end{table*}

	\subsection{Experimental setup}
	The experiments were conducted on a Linux system using Python
	. Our method is implemented using the Tensorflow library. Each GPU based experiment was conducted on an Nvidia 1080TI GPU. When the abstract is available, a paper (node) attribute is given as a concatenation of both the title and abstract else only the title is used. Each citation relationship (edge) attribute is given as the concatenation of all its citation contexts (i.e., sentences where the reference is mentioned in the citing paper). The paper and citation attributes are then converted to a vector by applying the latent semantic analysis (LSI) method on the document-term matrix features, resulting in 300-dimension features vectors. We complete the missing citation attributes with zero vectors and assume no missing paper attribute. In all the experiments, the attribute vector is normalized to unit norm. 
	
	For our proposed model, we performed a grid search over the length of walk $T=\{5, 10, 20, 40\}$ and the number of walks per node $M = \{1, 5,10,20\}$. For each neural network based model, we performed a grid search over the learning rate $lr = \{1e^{-2}, 5e^{-2}, 1e^{-3},$ $5e^{-4}, 1e^{-4} \}$ and hidden layer dimension $d = \{32,64,128\}$.  We performed the parameter grid search by training on the CoraL1 dataset with 10\% labeled samples. The best parameters per model from the grid search are then used in all experiments. 	
	The RAW models are trained for 30 epochs with a parameter set ($d=128, T=10$, $lr = 1e^{-4}$, $M = 5$ for training and $M=10$ for testing). The GCN and FastGCN models are trained for 200 epochs with $ lr = 1e^{-2}$ and $d = 64$ and $128$ respectively. The GraphSAGE and GAT models are trained for 20  and 100 epochs respectively with a parameter set of ($d= 128, lr=1e^{-2}$). The Planetoid models are trained for 5000 epochs with a parameter set of ($d=64, lr = 0.1$). 	
	We used the Scikit-Learn implementation of Linear SVM with default settings for embedding based evaluations. All experiment results reported in this paper are averaged from running on each dataset five times on random samples. For each experiment, we separate 30\% of the labeled data for testing. We then vary the number of labeled training data, with the remaining labeled samples assumed to be unlabeled (included in the set of unlabeled samples).  In all our experiments, we assume the graph to be undirected.
	
	\subsection{Comparison Methods}
	
	To evaluate the performance of our model, we compare RAW with several state-of-the-art semi-supervised graph-based methods using classification accuracy as the performance metric. We selected the most competitive baselines that are also publicly available online to avoid unfair evaluations due to faulty implementation. 
	The baseline methods are from different groups: 
	
	\textbf{Unsupervised embedding + classifier:} we generate embeddings using several unsupervised embedding methods, which we then give as input to the Linear SVM model for training and classification. The embedding methods include: Node2Vec \cite{grover2016node2vec}, DeepWalk \cite{perozzi2014deepwalk}, Latent Semantic Analysis \cite{deerwester1990indexing}, and TADW \cite{yang2015network}. 
	
	\textbf{Semi-supervised learning on graph:}  we selected the most popular models including Planetoid \cite{yang2016revisiting}, and GCN \cite{gcn}.  
	
	\textbf{Supervised learning on graph:} in the inductive setting, we evaluate against several variants of FastGCN \cite{chen2018fastgcn} and GraphSAGE \cite{hamilton2017inductive} which are supervised learning models for inductive node classification. Note, however, that our proposed method works in a semi-supervised manner in both the inductive and transductive settings.
	
	\textbf{Semi-supervised learning on graph with attention:} like our RAW,  GAT \cite{velickovic2017graph} and AGNN \cite{thekumparampil2018attention} employed attention mechanism when aggregating the neighbors. Note that we only show the results of GAT   due to the poor performance of AGNN on our datasets.
	

	\subsection{Results}
	
	\subsubsection{\textbf{Transductive}}
	Table \ref{experiment_result} shows the classification performance of our proposed model and other state-of-the-art models. Our proposed model exhibited similar performance compared with GCN in the transductive settings, but it outperforms all other baseline methods in all settings. 
	For the GAT models, we use the sparse version (SpGAT) as the original implementation gave an out of memory error (OOM) on even the CoraL1 with 10\% labeled samples. We could only evaluate the GCN and GAT on the Cora datasets as we got out of memory error when applying them to  the other  large datasets due to the dense LSI vectors. 
	We will evaluate the scalability of these methods and show the memory usage analysis in section \ref{sec:para_memory}.
	In summary, RAW is usable on large-scale graphs and produce the best node classification results, with no significant difference to GCN, but more efficient than GCN. 
	
	\subsubsection{\textbf{Inductive}}
	Table \ref{experiment_result} also shows the comparison of RAW and other 
	inductive models. RAW outperforms all the baseline methods in all settings. In the inductive setting, the testing nodes are removed from the training graph and thus are not seen during training. The agent learns the optimal policy for the graph walk during training that will be generalized to unseen nodes. The test nodes are only added to the graph during testing. The agent (guided by the policy learned after training), starts a walk from the added nodes to learn the embedding for the new nodes. We compare RAW against GraphSAGE, FastGCN and Planetoid inductive model. GraphSAGE and FastGCN are supervised learning algorithms and thus do not use the unlabeled and test nodes during training. The superior performance of RAW shows that walks starting from the new nodes guided by the learned policy aggregated the most useful information for classifying the starting node (the target to classify). 
	

	\begin{figure}[t]
		\includegraphics[width=0.40\textwidth]{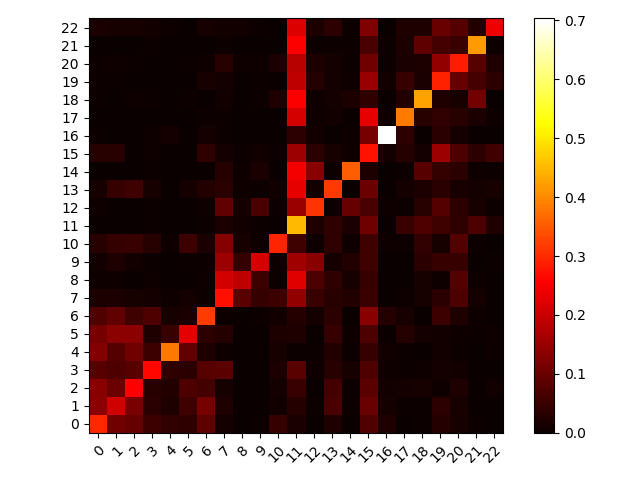}\vspace{-0.2cm}
		\caption{ 
			A heatmap whose $d$-th column demonstrates the RAW agent starting from nodes with label $d$ moved to nodes with what label distribution. The visiting frequency rate is shown in color. Brighter color indicates more visits. 
		}
		\label{class_rel}
	\end{figure}
	

	\subsubsection{\textbf{Trajectory Analysis}}
	Furthermore, 
	we analyze the returned walk trajectory from RAW. This study is performed on the CoraIDA dataset with $T=30$. 
	We extract the trajectories learned for nodes in each class, and then get the distribution of labels for all nodes visited on these trajectories.
	The 23 columns in the whole heatmap plot correspond to 23 class labels given in Table \ref{class_map}. 
	From figure \ref{class_rel}, we can see that the walk sequences for each class mostly visit the nodes in the same class as the target class (the light squares on diagonal). This verifies that RAW agent tends to walk to  nodes in the same class for accomplishing the classification task. It is worth mentioning that  RAW agent has no information about label when walking, neither   the target label (label of the starting node), nor   the label of neighboring nodes.

	\begin{table}[h]
		\caption{Class label IDs of the CoraIDA dataset} \vspace{-0.2cm}
		\label{experiment_result}
		\scriptsize 
		\begin{tabular}{|l|l|l|l|}
			\hline
			Class               & ID & Class          & ID \\ \hline
			DB/Object Oriented  & 0  & AI/Machine Learning               & 11 \\ \hline
			DB/Query Evaluation & 1  & AI/NLP         & 12 \\ \hline
			DB/Relational       & 2  & AI/Data Mining & 13 \\ \hline
			DB/Temporal         & 3  & AI/Speech      & 14 \\ \hline
			DB/Concurrency      & 4  & AI/Knowledge Representation       & 15 \\ \hline
			DB/Performance      & 5  & AI/Theorem Proving                & 16 \\ \hline
			DB/Deductive        & 6  & AI/Games and Search               & 17 \\ \hline
			IR/Retrieval        & 7  & AI/Vision and Pattern Recognition & 18 \\ \hline
			IR/Filtering        & 8  & AI/Planning    & 19 \\ \hline
			IR/Extraction       & 9  & AI/Agents      & 20 \\ \hline
			IR/Digital Library  & 10 & AI/Robotics    & 21 \\ \hline
			&    & AI/Expert Systems                 & 22 \\ \hline
		\end{tabular}
		
		\label{class_map}
	\end{table}

	More importantly,   we observe in Figure \ref{class_rel} the relationship between the classes (note again RAW agent moves without any label information). For instance,   we can observe that papers under some topics in a research field tend to visit other papers in the same research field more often.  \emph{Database} papers (with label 0-6) form a block in the left-bottom  corner. The other two blocks, although not obvious but observable, correspond to \emph{information retrial} and \emph{artificial intelligence}.
	Figure  \ref{class_rel} also highlights the important topics. We can notice the influence of the \emph{Machine Learning} class on the \emph{Artificial Intelligence} and \emph{Information Retrieval} community. This influence is shown by the ratio of times the walk sequence of nodes in each class under \emph{AI} visits the \emph{machine learning} nodes. It is interpretable as an individual usually needs to read some machine learning papers/books to understand these topics better. We also notice the versatility of the classes. We see the walk sequence of the class \emph{Theorem Proving} mostly visit nodes in its class. This result shows that the research area is quite narrow while \emph{Machine Learning} and \emph{Knowledge representation} are broader topics and therefore, more versatile.  
	
\begin{figure}[]
		\centering
		\begin{subfigure}[b]{0.25\textwidth}
			\includegraphics[width=\textwidth]{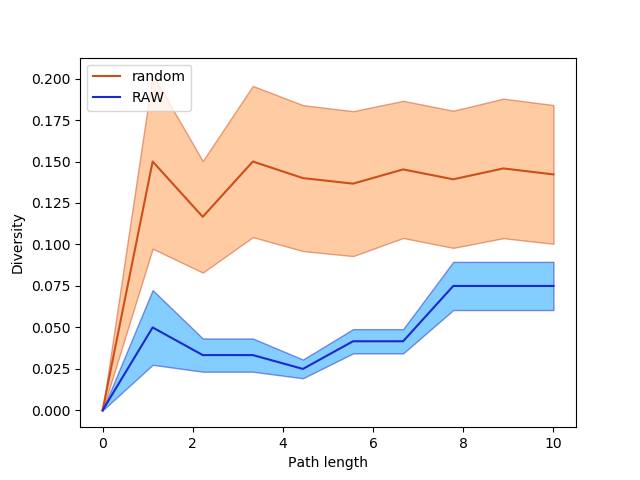}
			\caption{}
			\label{fig:walk1}
		\end{subfigure}
		~ 
		\begin{subfigure}[b]{0.25\textwidth}
			\includegraphics[width=\textwidth]{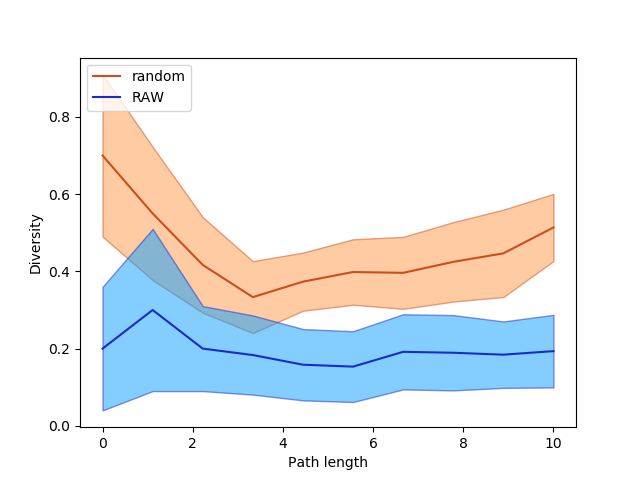}
			\caption{}
			\label{fig:walk2}
		\end{subfigure}
		\caption{The mean and variance of the \emph{path label diversity} defined in Eq.(\ref{eq:diversity}), measured on ten paths starting from two randomly sampled paper,   (a)  in class  \textit{``Concurrency"} and (b) in class \textit{``Vision and pattern recognition"}}\label{fig:walks}
	\end{figure}

	To further analyze the performance of RAW,   we compare the path made by RAW and  random walk. By setting $T=10$, we obtain a set of trajectories returned by RAW, and another set of trajectories by random walk starting from the same node.  For each trajectory, we calculate the \emph{path label diversity}  for each walking step $t$:
	\begin{equation} \label{eq:diversity}
	\delta_t = 1 - \frac{\Sigma_{i=0}^t \mathbbm{1}(l_i = l_0)}{t},
	\end{equation}
	where $l_0$ is the  label of the starting node,   $l_i$ is the label of the $i$-th node on the path, and $\mathbbm{1}()$ is the indicator function. The value $\delta_t$ is low (to 0) when nodes on the path have the same label as the starting node, indicating  that the agent learned  to explore neighboring nodes with the same label as the target label. Note that the agent has no label knowledge during the walk. Figure \ref{fig:walks} shows the mean and variance of the \emph{path label diversity} when starting at two different selected nodes. We can see that RAW agent walks with a much lower diversity than random walk. 
	
	
	\begin{figure}[]
		\includegraphics[width=0.40\textwidth]{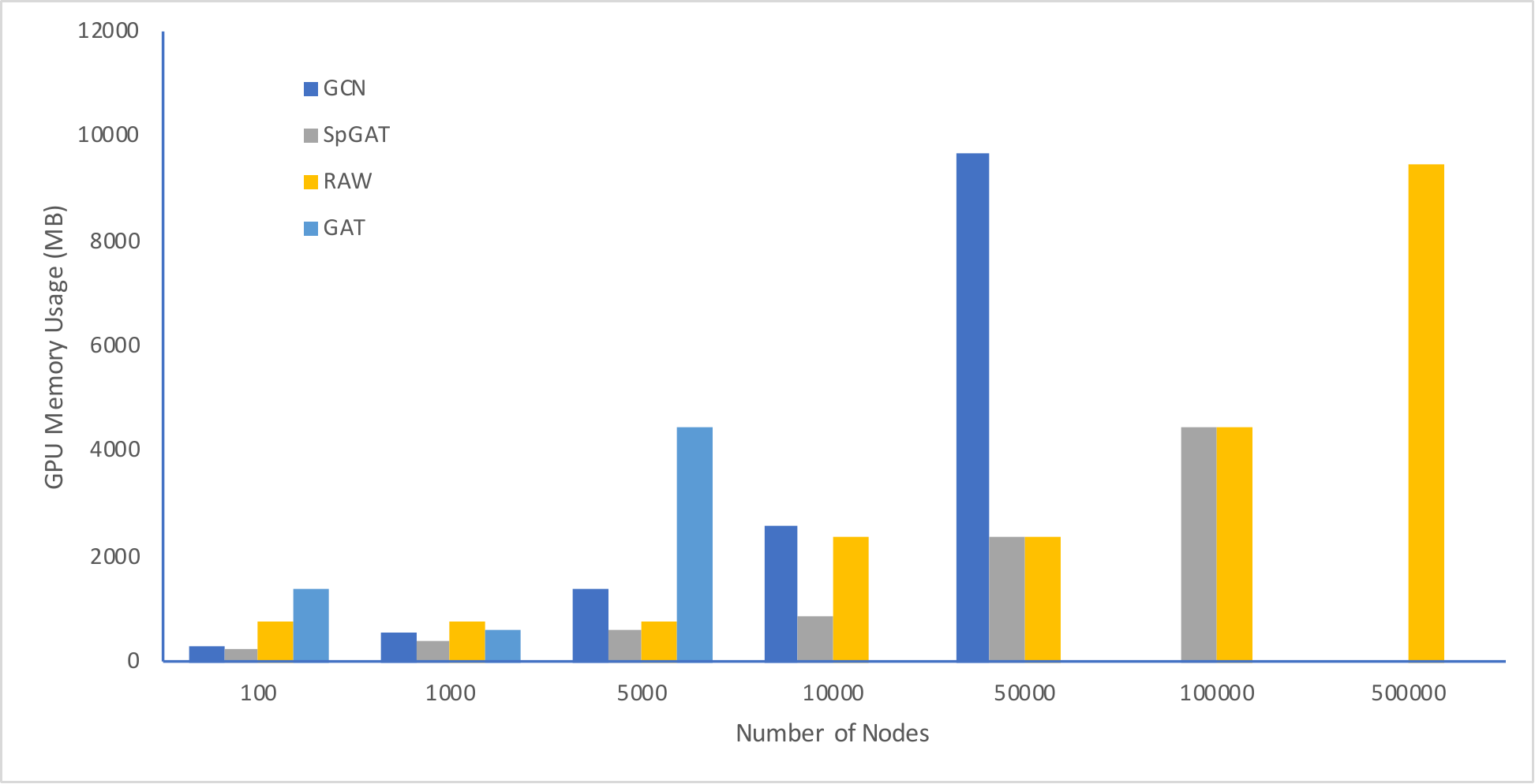}
		\caption{GPU memory usage. Missing bars indicate  an out of memory error (OOM).}
		\label{memory}
	\end{figure}

	\begin{figure}[]
		\includegraphics[width=0.30\textwidth]{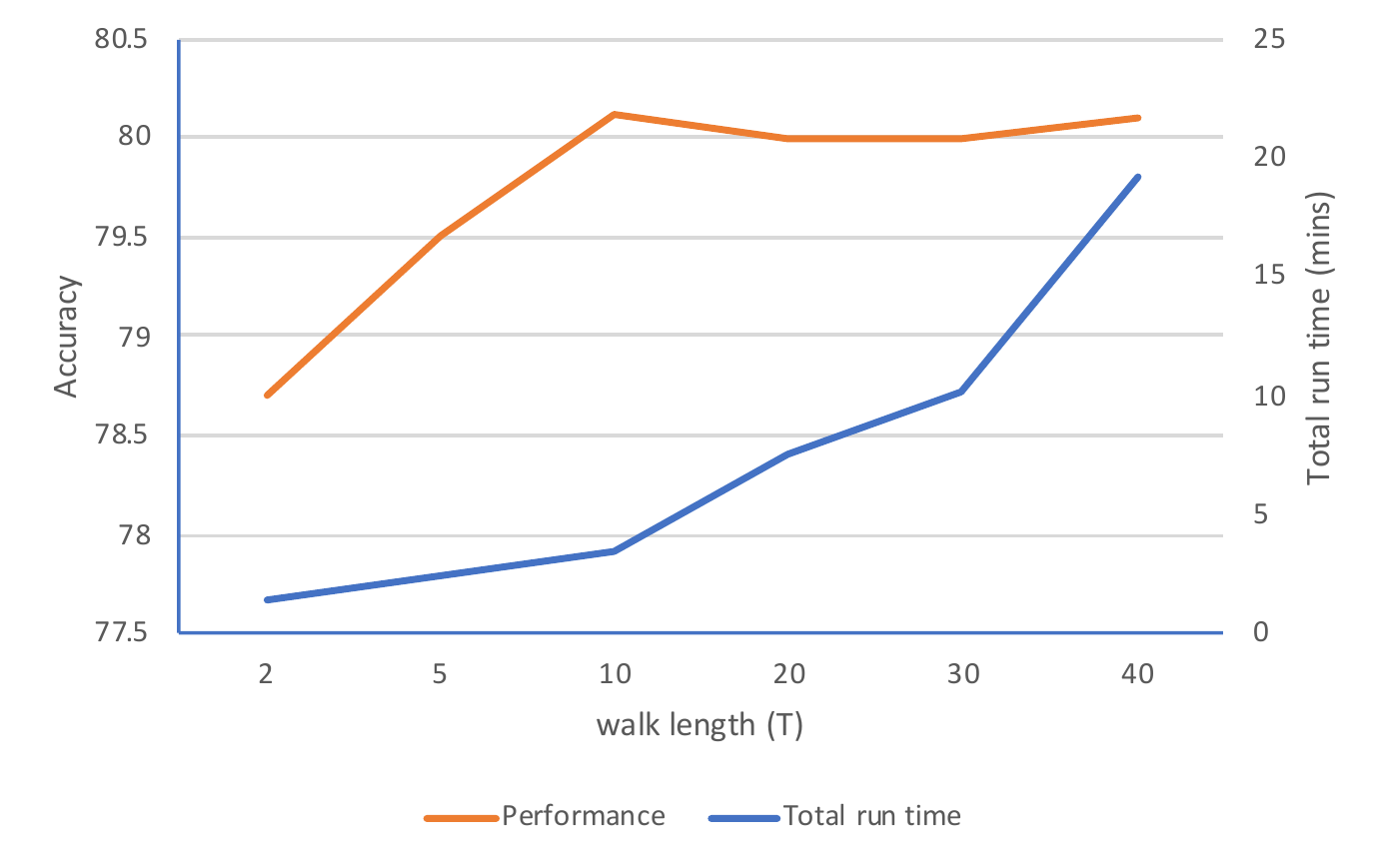}
		\caption{Effect of the walk length on the predictive performance and running time.}
		\label{parameter}
	\end{figure}
	\vspace{-0.1cm}
	
	\subsubsection{\textbf{Parameter and Memory Analysis}} \label{sec:para_memory}
	We study the effect of the walk length $T = \{2, 5, 10, 20, 30, 40\}$ on the performance of the model. We train the model on the CoraL1 dataset with 10\% training samples.  In Figure \ref{parameter}, it can be observed that the model already performs well after ten steps as there is no much improvement with an increase in the number of walks. Meanwhile, more  number of walks causes higher time cost. 
	
	Figure \ref{memory} shows the GPU memory utilization of our proposed model and several semi-supervised state-of-the-art transductive models. We randomly generated Erd\H{o}s-R\'{e}nyi graphs in size of 100, 1K, 5K, 10K, 50K, 100K and 500K (the number of nodes),  and  set the number of edges in each graph to be ten times the number of nodes. We randomly generate 300-dimension attributes for the nodes and edges. 
	We then measure the GPU memory consumption using the \textit{nvidia-smi} Linux command on each graph and compare RAW with GCN, GAT and SpGAT. 
	GCN and GAT do not scale with the number of nodes and edges (shown with  zero bars in Figure \ref{memory} after the graph gets larger than 50K). 
	
	\section{Case Study}
	\begin{figure}[]
		\centering
		\includegraphics[width=0.4\textwidth]{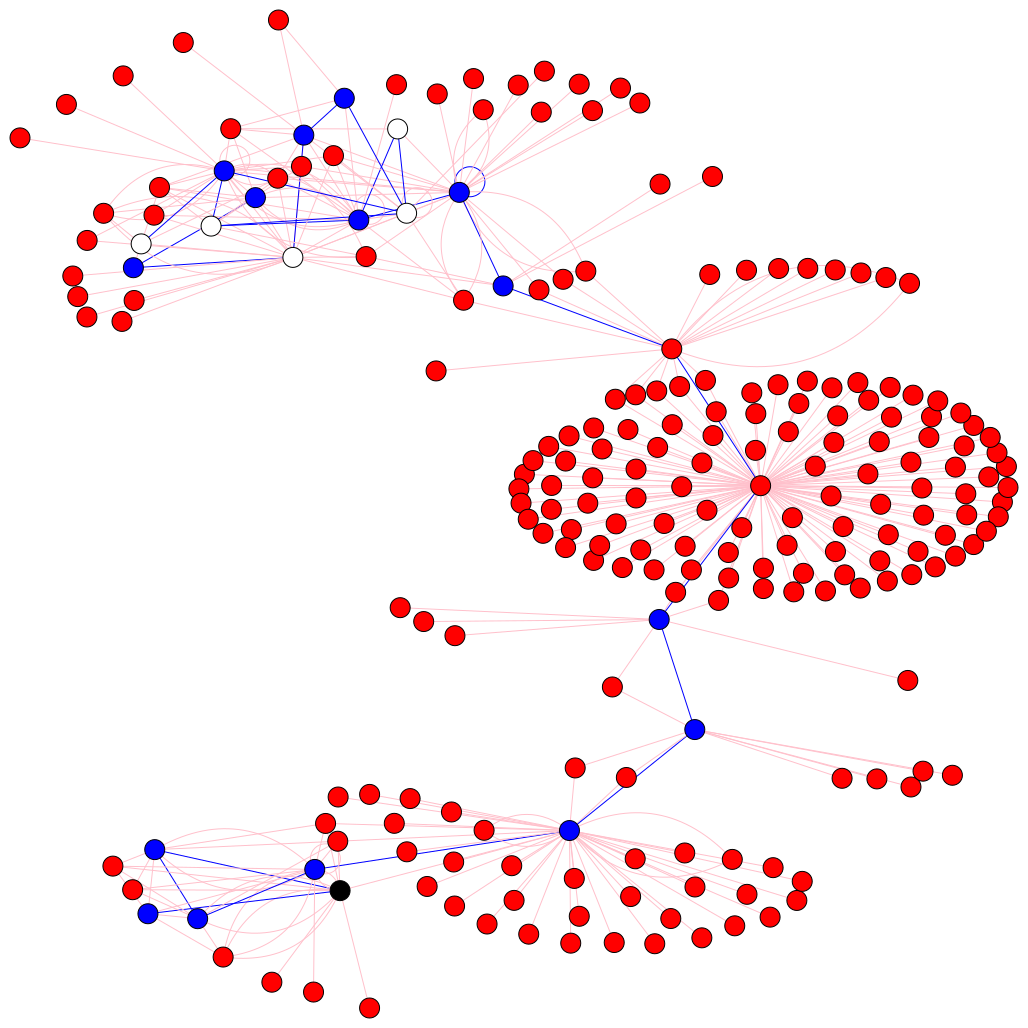}
		\caption{Case study of a sampled walk trajectory, starting from the node (black node) of a paper entitled ``\textit{A Critique of Structure from motion Algorithm}'' classified as ``Vision and Pattern Recognition''  on a subgraph of the CoraIDA graph.} 
		\label{graph_path}
	\end{figure}
	\vspace{-0.1cm}
	In this section, we train the RAW model on the CoraIDA dataset with a trajectory length of $T=30$ and present a case study of a sampled walk trajectory of a paper. Figure \ref{graph_path} shows the walk sequence extracted from a paper entitled ``\textit{A Critique of Structure from motion Algorithm}'' classified as ``Vision and Pattern Recognition''  on a subgraph of the CoraIDA graph. The thickness of the edges signifies the ratio of times the edge was traversed during the walk. The color of the nodes signifies the class relationship of the node to the target class. The blue color signifies that a node has the same label as the target label, the red signifies that a node has a label different to the target node, and the black color signifies the start node.  Note that the target class is the class of the start node. 
	
	We see from Figure \ref{graph_path} that {\bf the agent can selectively make decisions to visit nodes with the same labels as that of the start node}. The agent also visits unlabeled nodes (white nodes in the right-bottom corner). We observe that even though the labels are unknown, the visited unlabeled nodes work on similar topics as the start paper, e.g., entitled   ``\textit{new statistical models for randoms-precorrected pet scans}'', ``\textit{fast monotonic algorithms for transmission tomography}'',  etc. 

	\section{Conclusion}
	In this paper, we propose to address the semi-supervised node classification problem in attributed networks by letting an agent choose the most relevant nodes in a recurrent walk framework. 
	The decision of where to visit is determined by considering the previous visiting history, the current node content, node content of node one-hop neighbors,  and the edge content between the current node and its linked neighbors. The accumulated information from the nodes in the sequence is finally used for classification. 
	We show by several experiments and analysis that the proposed model outperforms several state-of-the-art methods in both transductive and inductive settings. The analysis of the obtained walk sequences also confirms that our model selects the most relevant nodes to visit and thus leads to higher classification accuracy than other methods.
	
	\section{Acknowledgment}
	This work is supported by the King Abdullah University of Science and Technology (KAUST), Saudi Arabia
%
	
	%
	\bibliographystyle{ACM-Reference-Format}
	\bibliography{bibliography}

\end{document}